\documentclass[lettersize,journal]{IEEEtran}
\usepackage{amsmath,amsfonts}
\usepackage{algorithmic}
\usepackage{algorithm}
\usepackage{array}
\usepackage[caption=false,font=normalsize,labelfont=sf,textfont=sf]{subfig}
\usepackage{url}
\usepackage{graphicx}
\usepackage{cite}
\usepackage{booktabs}
\usepackage{multirow}
\usepackage{color}
\usepackage{ulem}
\usepackage{amssymb}
\newcommand{\settablefont}{\fontsize{6}{12}\selectfont}
\newcommand{\eg}{\textit{e.g.}}
\newcommand{\ie}{\textit{i.e.}}

\newcommand{\etal}{\textit{et al. }}

\usepackage[hidelinks,hypertexnames=false, colorlinks]{hyperref}
\usepackage[table,xcdraw]{xcolor}
\usepackage{scalerel}
\usepackage{tikz}
\usetikzlibrary{svg.path}
\definecolor{orcidlogocol}{HTML}{A6CE39}
\tikzset{
	orcidlogo/.pic={
		\fill[orcidlogocol] svg{M256,128c0,70.7-57.3,128-128,128C57.3,256,0,198.7,0,128C0,57.3,57.3,0,128,0C198.7,0,256,57.3,256,128z};
		\fill[white] svg{M86.3,186.2H70.9V79.1h15.4v48.4V186.2z}
		svg{M108.9,79.1h41.6c39.6,0,57,28.3,57,53.6c0,27.5-21.5,53.6-56.8,53.6h-41.8V79.1z M124.3,172.4h24.5c34.9,0,42.9-26.5,42.9-39.7c0-21.5-13.7-39.7-43.7-39.7h-23.7V172.4z}
		svg{M88.7,56.8c0,5.5-4.5,10.1-10.1,10.1c-5.6,0-10.1-4.6-10.1-10.1c0-5.6,4.5-10.1,10.1-10.1C84.2,46.7,88.7,51.3,88.7,56.8z};
	}
}
\newcommand\orcidicon[1]{\href{https://orcid.org/#1}{\mbox{\scalerel*{
				\begin{tikzpicture}[yscale=-1,transform shape]
					\pic{orcidlogo};
				\end{tikzpicture}
			}{|}}}}
\begin{document}
	
\title{SCIPaD: Incorporating Spatial Clues into Unsupervised Pose-Depth Joint Learning}
\author{Yi Feng$^{\orcidicon{0009-0005-4885-0850}\,}$, 
	Zizhan Guo$^{\orcidicon{0009-0003-7360-2192}\,}$, 
	Qijun Chen$^{\orcidicon{0000-0001-5644-1188}\,}$,~\IEEEmembership{Senior Member,~IEEE}, 
	and Rui Fan$^{\orcidicon{0000-0003-2593-6596}\,}$,~\IEEEmembership{Senior Member,~IEEE}
\thanks{
This research was supported by the Science and Technology Commission of Shanghai Municipal under Grant 22511104500, the National Natural Science Foundation of China under Grant 62233013, the Fundamental Research Funds for the Central Universities, and the Xiaomi Young Talents Program. (\emph{Corresponding author: Rui Fan})}
\thanks{Yi Feng, Zizhan Guo, Qijun Chen, and Rui Fan are with the Department of Control Science \& Engineering, the College of Electronics \& Information Engineering, Shanghai Research Institute for Intelligent Autonomous Systems, Shanghai Institute of Intelligent Science and Technology, the State Key Laboratory of Intelligent Autonomous Systems, and Frontiers Science Center for Intelligent Autonomous Systems, Tongji University, Shanghai 201804, China (e-mails: fengyi@ieee.org, \{2052110, qjchen\}@tongji.edu.cn, rui.fan@ieee.org).}
}

\maketitle
	
\begin{abstract}
Unsupervised monocular depth estimation frameworks have shown promising performance in autonomous driving. However, existing solutions primarily rely on a simple convolutional neural network for ego-motion recovery, which struggles to estimate precise camera poses in dynamic, complicated real-world scenarios. These inaccurately estimated camera poses can inevitably deteriorate the photometric reconstruction and mislead the depth estimation networks with wrong supervisory signals. In this article, we introduce SCIPaD, a novel approach that incorporates spatial clues for unsupervised depth-pose joint learning. Specifically, a confidence-aware feature flow estimator is proposed to acquire 2D feature positional translations and their associated confidence levels. Meanwhile, we introduce a positional clue aggregator, which integrates pseudo 3D point clouds from DepthNet and 2D feature flows into homogeneous positional representations. Finally, a hierarchical positional embedding injector is proposed to selectively inject spatial clues into semantic features for robust camera pose decoding. Extensive experiments and analyses demonstrate the superior performance of our model compared to other state-of-the-art methods. Remarkably, SCIPaD achieves a reduction of 22.2\% in average translation error and 34.8\% in average angular error for camera pose estimation task on the KITTI Odometry dataset. Our source code is available at \url{https://mias.group/SCIPaD}.	
\end{abstract}

		
\section{Introduction}
\IEEEPARstart{A}{UTONOMOUS} vehicles are gradually becoming an integral part of our daily lives \cite{li2024roadformer}. Monocular depth estimation plays a crucial role in the perception systems of autonomous vehicles, as it directly enables agents to perform scene parsing \cite{fan2020sne, fan2019road, fan2021graph}, self-localization \cite{feng2023freespace}, and scene reconstruction \cite{digging2019godard, fan2018road, feng2023d2nt}. Early works \cite{depth2014eigen, learning2016liu} solve the monocular depth estimation problem via supervised learning, which requires precise, extensive depth ground truth, typically acquired using additional cameras or LiDARs \cite{huang2024online, zhao2023dive}. However, it is time-consuming and labor-intensive to gather large-scale depth data from the real world, and the specific data distribution can restrict the generalizability of the network to new, unseen scenarios. To address these limitations, self-supervised methods have emerged as favorable alternatives, which generate supervisory signals from stereo pairs or monocular videos to jointly estimate depth and camera pose (also referred to as ego-motion), thereby eliminating the necessity for ground-truth depth acquisition.

\begin{figure}
\centering
\includegraphics[width=0.99\linewidth]{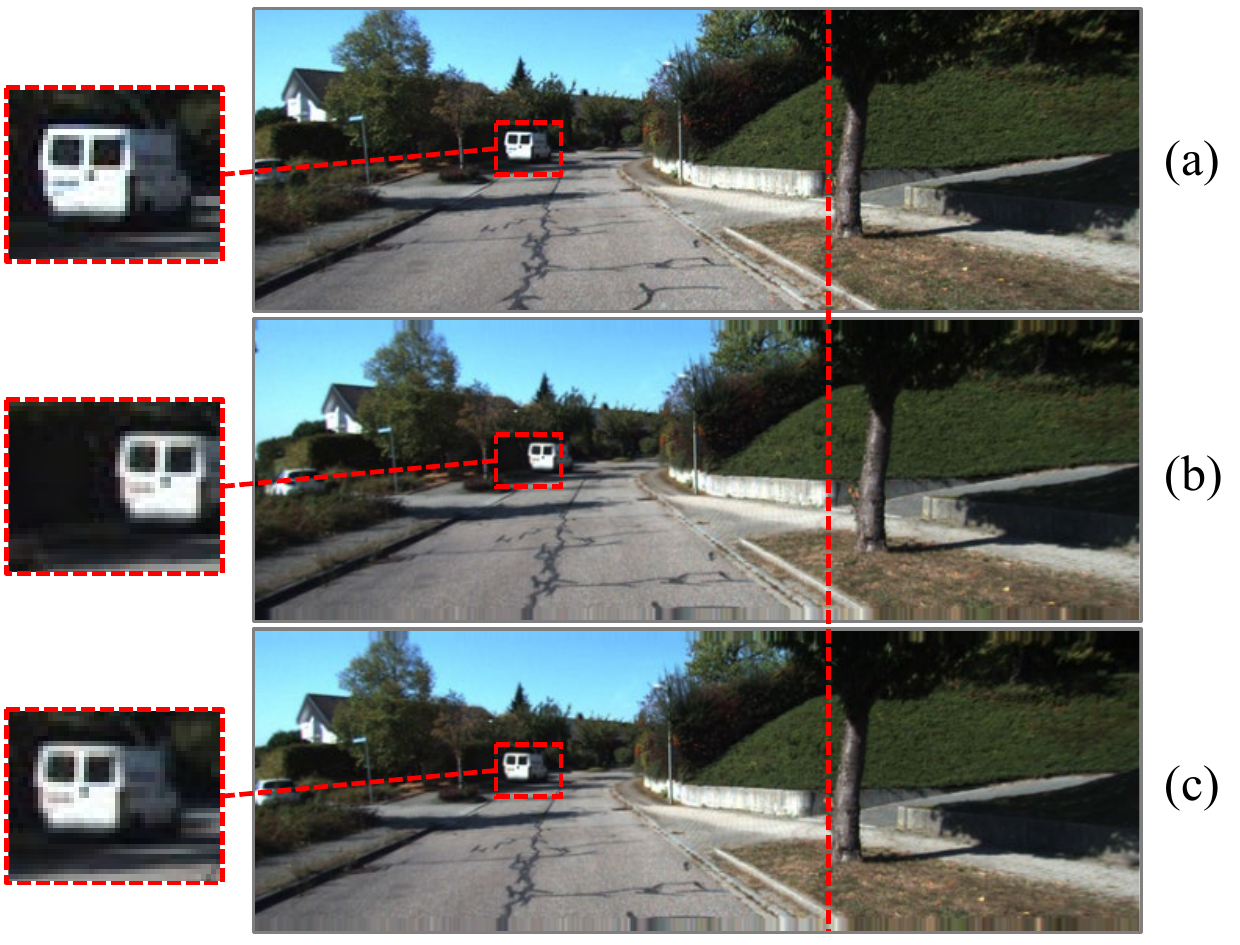}
\caption{Photometric reconstruction comparison between SQLdepth \cite{wang2024sqldepth} and our proposed SCIPaD: (a) the original target frame; (b) the image reconstructed using SQLdepth; (c) the image reconstructed using SCIPaD. Red boxes and vertical lines are added for visual comparison. Our method demonstrates superior performance in camera pose and depth joint estimation.}
\label{fig.recons}
\end{figure}

In recent years, self-supervised monocular depth estimation has attracted considerable attention, with ongoing efforts aimed at addressing corner cases in complex and dynamic environments. These self-supervised methods typically utilize a photometric reconstruction loss as the supervisory signal, which can be affected by the precision of both depth predictions and camera pose estimations. Nevertheless, existing frameworks tend to prioritize depth estimation accuracy while often neglecting the accuracy of pose estimation, which is currently generated by a shallow convolutional neural network (CNN)-based PoseNet \cite{posenet2015kendall}.

The adopted PoseNet architecture suffers from three key limitations: First, the acquisition of ego-motion is strongly related to positional clues and geometrical constraints. Classical visual odometry algorithms \cite{campos2021orb, qin2018vins} utilize the positions of matched keypoints to recover camera poses, leveraging epipolar geometry and perspective-n-point. However, current learning-based frameworks take concatenated RGB frames as input and use encoder-decoder structures for 6-DoF pose regression. Despite their computational efficiency, these structures lack geometrical reasoning and positional encoding abilities. Second, the depth estimation network (hereafter referred to as DepthNet) provides valuable priors of 3D spatial layout. Existing methods fail to utilize the inherent knowledge of spatial layout to enhance the interpretability and precision of ego-motion estimations, further leading to sub-optimal performance in depth-pose joint estimation. Third, the PoseNet backbone is typically pretrained on image classification datasets such as ImageNet \cite{imagenet2009deng}, which excels in extracting semantic information rather than modeling geometrical correlations \cite{jia2024feature}. Moreover, these networks typically use the deepest semantic features for camera pose regression, while discarding the shallower ones which could potentially contain informative spatial clues and positional correlations. 
	
To address the aforementioned issues, we first propose a confidence-aware feature flow estimator (\textbf{CAFFE}) to calculate and adjust dense feature correspondences with the consideration of pixel-wise confidence levels. This module explicitly extracts abundant positional clues regarding 2D feature translations, which provides strong constraints for ego-motion recovery. Additionally, we argue that the output of DepthNet offers valuable priors of 3D geometrical layout, which may enhance the consistency between camera pose and depth predictions. Therefore, we introduce a positional clue aggregator (\textbf{PCA}), which incorporates 2D feature correspondences from CAFFE and 3D spatial layout from DepthNet into a homogeneous positional embedding space. Finally, it is observed that the deeper layers of PoseNet encode richer semantic information, whereas the shallower layers emphasize precise spatial cues. Building upon this insight, we introduce a hierarchical positional embedding injector (\textbf{HPEI}), which selectively incorporates positional embeddings into semantic clues with learnable gates. As demonstrated in the experiments, the performance of our model is enhanced by the adaptive integration of both semantic and spatial information. 
	
In conclusion, we propose \uline{\textbf{SCIPaD}, a \textbf{S}patial \textbf{C}lue-\textbf{I}ncorporated \textbf{P}ose \textbf{a}nd \textbf{D}epth joint learning framework}, which integrates all aforementioned innovative components. It demonstrates superior performance in pose and depth joint estimation compared with other state-of-the-art (SoTA) methods, especially in ego-motion recovery. As illustrated in Fig. \ref{fig.recons}, we compare the photometric reconstruction performance between SQLdepth \cite{wang2024sqldepth} and our proposed SCIPaD.  It can be seen that current SoTA method often fails to produce precise camera poses, which further misleads the photometric consistency-based supervisory signals. 

In summary, we make the following contributions: 
\begin{enumerate}
\item We propose SCIPaD, a novel monocular depth-pose joint learning framework with spatial clues incorporated, achieving SoTA performance across multiple public datasets. 
		
\item We introduce a CAFFE for calculating and reweighting dense feature correspondences, in which a novel 2D soft argmax function is utilized for differentiable dense feature matching.

\item We develop a PCA, which incorporates 2D feature flow and 3D spatial layout into a homogeneous representation of positional clues, ensuring comprehensive geometry encoding and proving highly effective in ego-motion estimation tasks.

\item We propose an HPEI, which selectively injects positional embeddings into semantic clues through a learnable gating mechanism, leading to improved performance in both depth and camera pose estimation. 	
\end{enumerate}
	   
The remainder of this article is structured as follows. Sect. \ref{sec.related_work} presents an overview of the SoTA monocular depth estimation and camera pose estimation methods. In Sect. \ref{sec.method}, we introduce the proposed SCIPaD framework for depth-pose joint learning. In Sect. \ref{sec.exp}, we present the experimental results across several public datasets. Sect. \ref{sec.conclusion} provides a detailed discussion and concludes this article.

\section{Related Work}
\label{sec.related_work}
Depth estimation is a fundamental task in computer vision and robotics \cite{wu2024s}, involving the use of RGB images to generate dense depth predictions. %Each pixel in the resulting depth map indicates the distance between the observed object and the camera's focal point \cite{han2023self}. 
%Conventionally, depth estimation tasks are classified into three types: monocular, binocular, and multi-view \cite{ming2021deep}. Among them, 
In recent years, monocular depth estimation has garnered significant attention due to its wide applications in autonomous driving \cite{xue2020toward, 3d2020guizilini} and virtual reality \cite{el2019survey, li2020unsupervised}. 
%However, monocular depth estimation is an ill-posed problem, as a single 2D image can correspond to multiple possible 3D scenes at different scales \cite{depth2014eigen}. 
To date, deep learning-based approaches for this task generally fall into two categories: supervised and unsupervised \cite{zhang2024dcpi}.

\subsection{Supervised Monocular Depth Estimation}
The supervised learning approach for this task requires pixel-level depth ground truth during the training phase. Eigen \etal \cite{depth2014eigen} first introduced a deep learning model with a coarse-to-fine architecture to predict depth maps. Subsequent research has focused on employing more complex network structures \cite{estimating2018cao, learning2016liu, monocular2021song} or loss functions \cite{enforcing2019yin, deep2018fu} to enhance depth estimation performance.

Cao \etal \cite{estimating2018cao} and Fu \etal \cite{deep2018fu} reformulated the depth regression problem as a classification task, aiming to predict depth ranges rather than exact depth values.
Lee \etal \cite{big2021lee} introduced multi-scale guidance layers to establish the connections between intermediate layer features and the final depth map. 
Bhat \etal \cite{adabins2021bhat} proposed AdaBins, which adapts bin sizes based on image content, improving the adaptability of depth prediction.
Yang \etal \cite{transformerbased2021yang} developed a Vision Transformer-based architecture to capture long-range correlations in depth estimation. Recently, Depth Anything \cite{depth2024yang} achieved impressive results by fully unleashing the potential of foundation models. It first reproduced a MiDaS-based \cite{birklMiDaSV3Model2023} teacher model with DINOv2 \cite{dinov22023oquab} pretrained weights, and then used these teacher predictions as pseudo labels to train a student model on extensive unlabeled data (62M). 

Despite the promising results of these approaches, the requirement for substantial amounts of ground-truth depth labels in supervised training can be costly and limits the widespread adoption of these methods.

\subsection{Unsupervised Monocular Depth Estimation}

In the absence of ground-truth depth data, an unsupervised framework aims to train depth estimation models using image reconstruction as a supervisory signal. Garg \etal \cite{unsupervised2016garg} approached depth estimation as a novel view synthesis problem, minimizing the photometric loss between an input left image and the synthesized right image. Building upon this, Godard \etal \cite{unsupervised2017godard} enhanced accuracy by introducing a left-right disparity consistency loss.

In addition to using stereo pairs, unsupervised methods can learn depth estimation from monocular video frames. Zhou \etal \cite{unsupervised2017zhou} developed a network that jointly estimates depth maps and camera poses between sequential frames. To tackle challenges such as dynamic scenes and occlusions, other researchers have explored multi-task learning, which incorporates additional tasks like optical flow estimation \cite{geonet2018yin, competitive2019ranjan} and semantic segmentation \cite{selfsupervised2023chen, finegrained2021jung}. Furthermore, some researchers have introduced additional constraints, such as uncertainty estimation \cite{uncertainty2020poggi, d3vo2020yang}, to improve the robustness and accuracy of the models. Godard \etal \cite{digging2019godard} proposed Monodepth2, which leverages a minimum reprojection loss to mitigate occlusion issues and an automasking loss to filter out moving objects with velocities similar to the camera. Watson \etal \cite{temporal2021watson} introduced ManyDepth, which utilizes multiple frames at test time and leverages geometric constraints through cost volume construction, achieving superior performance.

Despite these advancements, all current unsupervised methods still rely on simple CNN-based PoseNet architectures for camera pose estimation, which often exhibit limited generalizabilities and consequently result in suboptimal depth reconstruction performance.

\subsection{Camera Pose Estimation}

Structure from motion (SfM) is widely recognized as a benchmark technique for 3D reconstruction and camera trajectory recovery from videos and image collections. Numerous studies \cite{undeepvo2018li, deep2020jau, better2020zhao, unsuperpoint2019christiansen} have endeavored to integrate neural networks into the SfM pipeline, fully leveraging the geometric priors learned from training data. 
Jau \etal \cite{deep2020jau} designed an end-to-end network for 6-DoF camera pose estimation. However, their method relies on highly accurate ground-truth camera poses, which are often unavailable or difficult to acquire. 
Li \etal \cite{undeepvo2018li} proposed UnDeepVO, an unsupervised visual odometry framework, capable of recovering absolute scales using monocular videos and stereo image pairs. 
Tang \etal \cite{selfsupervised2021tang} improved camera pose estimation performance by combining appearance and geometric matching through a differentiable SfM module. 
Bian \etal \cite{unsupervised2021bian} introduced a geometry consistency loss to enforce the scale-consistent depth learning. 

\begin{figure*}[!t]
    \centering
    \includegraphics[width=0.96\textwidth]{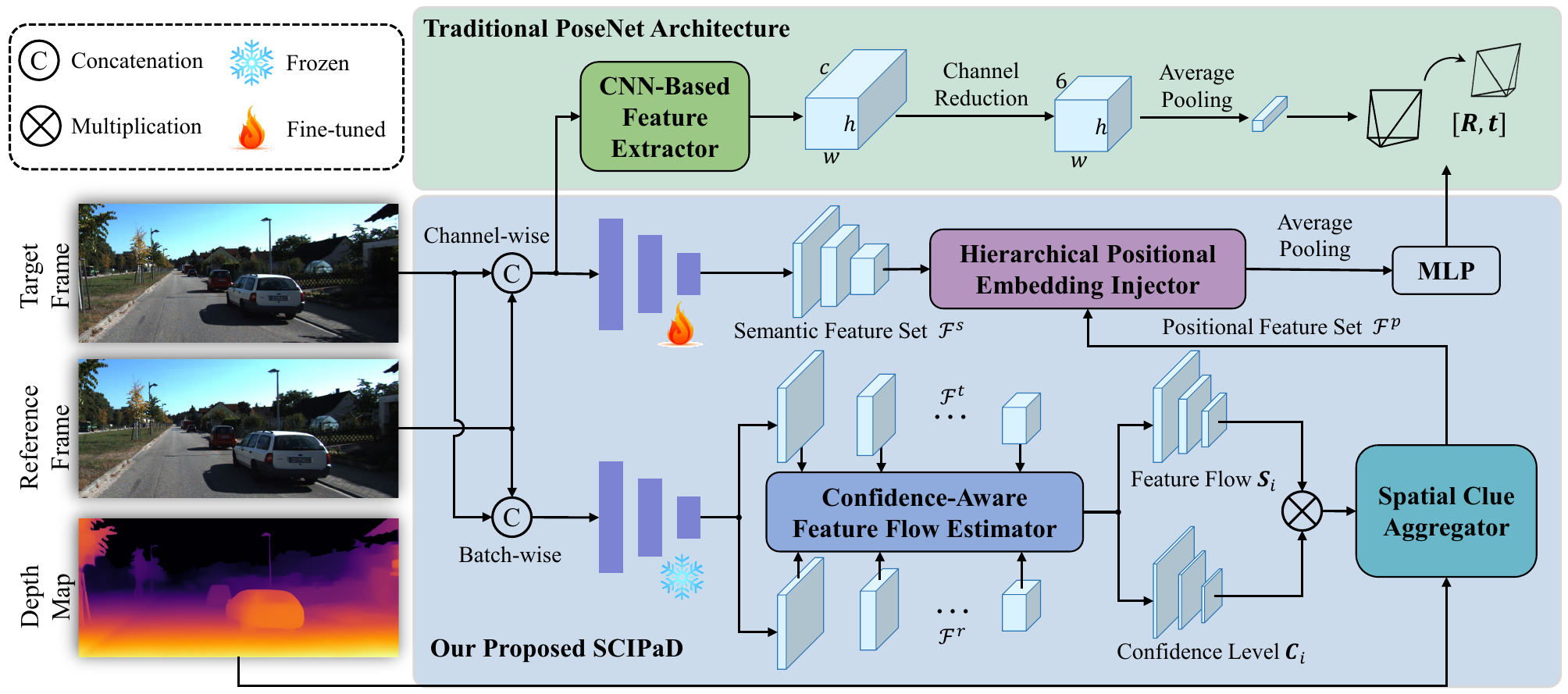}
    \caption{An illustration of our proposed SCIPaD framework. Compared with the traditional PoseNet \cite{posenet2015kendall} architecture, it comprises three main parts: (1) a confidence-aware feature flow estimator, (2) a spatial clue aggregator, and (3) a hierarchical positional embedding injector.}
    \label{fig.posenet}
\end{figure*}

Nonetheless, the pose estimation networks of these unsupervised SoTA methods heavily rely on a basic CNN-based architecture like PoseNet \cite{posenet2015kendall}. As shown in Fig. \ref{fig.posenet}, the model processes channel-wise concatenated monocular video frames, which are then passed through several convolutional layers for channel reduction, followed by an average pooling layer to produce a tensor of shape $1\times6$. This tensor, representing a combination of three Euler angles and three translational components, lacks interpretability for geometric modeling and robustness in scenarios involving moving objects.

To fully exploit the epipolar constraints, we explicitly encode the positional priors and incorporate depth information into camera pose estimation. Our system design and experimental findings provide novel insights that can significantly enhance the performance of both monocular depth estimation and camera pose estimation, providing valuable guidance for researchers in the field.

\section{Methodology}
\label{sec.method}
\subsection{Architecture Overview}

Previous studies such as \cite{wang2024sqldepth, liu2023towards, planedepth2023wang} adopt a CNN-based PoseNet \cite{posenet2015kendall} for camera pose estimation. Although the network architecture is effective and lightweight, it struggles to handle dynamic objects in the scene, which can significantly impair the performance of camera pose estimation, and further lead to failures in maintaining photometric consistency constraints. Moreover, the network backbone (usually ResNet-18 \cite{he2016deep}) is pretrained for image classification, which excels at extracting semantic cues from images. However, camera pose estimation requires the utilization of spatial and geometrical information, including 2D positional translations and 3D point cloud layouts, which have not been considered in the current SoTA frameworks. SCIPaD is thus proposed to solve the problems aforementioned. 

As illustrated in Fig. \ref{fig.posenet}, for the input reference and target frames $\boldsymbol{I}{^r}, \boldsymbol{I}{^t} \in \mathbb{R}^{3\times H\times W}$, meaningful semantic information is first extracted from two separate branches: one branch processes the channel-wise concatenated $\boldsymbol{I}{^r}$ and $\boldsymbol{I}{^t}$ to explore the implicit correlations between the semantics and estimated camera pose, producing a hierarchical semantic feature set $\mathcal{F}^s=\{\boldsymbol{F}{_1^s},...,\boldsymbol{F}{_k^s}\}$, where $\boldsymbol{F}{_k^s} \in \mathbb{R}^{c\times \frac{H}{2^k}\times \frac{W}{2^k}}$ represents the semantic features at the $k$-th stage. The other branch, which remains frozen, processes the batch-wise concatenated frames and separates them to produce translation-equivariant feature sets $\mathcal{F}^r=\{\boldsymbol{F}{_1^r},...,\boldsymbol{F}{_k^r}\}$ and $\mathcal{F}^t=\{\boldsymbol{F}{_1^t},...,\boldsymbol{F}{_k^t}\}$ of $\boldsymbol{I}{^r}$ and $\boldsymbol{I}{^t}$, respectively. Subsequently, confidence-aware feature flow is acquired by calculating feature affinity using a differentiable 2D soft argmax function, which is then integrated with the 3D point cloud data obtained from DepthNet to derive positional features $\mathcal{F}^p=\{\boldsymbol{F}{_1^p},...,\boldsymbol{F}{_k^p}\}$. Finally, the embedded positional features are hierarchically injected into the semantic features for 6-DoF camera pose regression. 
 
The following subsections detail the confidence-aware feature flow estimator, the spatial clue aggregator, and the hierarchical positional embedding injector within the SCIPaD framework.

\subsection{Confidence-Aware Feature Flow Estimator}

\begin{figure}[!t]
    \centering
    \includegraphics[width=0.49\textwidth]{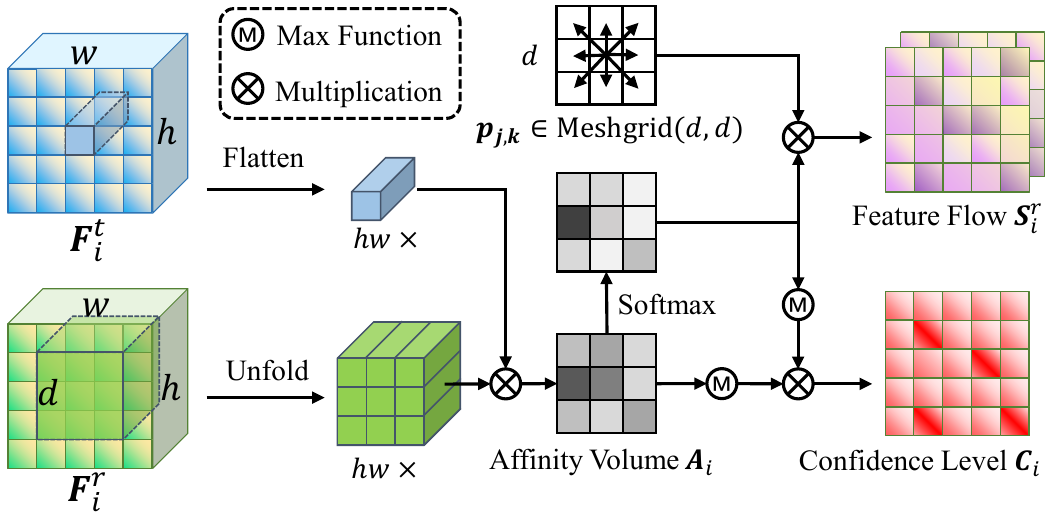}
    \caption{An illustration of our proposed confidence-aware feature flow estimator. It produces feature flow $\boldsymbol{S}^r_i$ and its confidence $\boldsymbol{C}_i$ through affinity volume construction and a differentiable 2D soft argmax function.}
    \label{fig.window_feature_affinity}
\end{figure}

As illustrated in Fig. \ref{fig.window_feature_affinity}, we take features $\boldsymbol{F}{_i^t}, \boldsymbol{F}{_i^r} \in \mathbb{R}^{c\times h\times w}$ from the target and reference frames at stage $i$ as an example, where $c$ is the feature channel dimension, and $(h,w) =\frac{(H,W)}{2^{i}}$ is the exponentially decreased feature resolution. 
We assume that the temporal intervals between successive frames are sufficiently short to ensure a relative constancy in object dimensions and to preserve the brightness constancy assumption \cite{horn1981determining} across corresponding points. Meanwhile, CNN-based models use a set of convolution kernels with shared weights for feature extraction, which exhibits a strong inductive bias of translation equivariance. Leveraging this property, we aim to find feature correspondences between the reference and target frames for explicit geometrical encoding. 

Unlike previous work \cite{deep2017zhu}, which primarily emphasizes feature flow generation across consecutive frames, our proposed CAFFE also produces pixel-wise confidence levels for reweighting the feature flow. We first normalize $\boldsymbol{F}{_i^r}$ to have a unit length along the channel dimension, and then extract sliding local blocks from the normalized features with window size $d$, resulting in the unfolded reference features $\tilde{\boldsymbol{F}}{^r_i} \in \mathbb{R}^{h\times w\times c\times d^2}$ as follows: 
\begin{equation}
    \tilde{\boldsymbol{F}}{^r_i} = 
    \operatorname{Reshape}(
    \operatorname{Unfold}(
    \frac{\boldsymbol{F}{_i^r}}{||\boldsymbol{F}{_i^r}||_2}
    )).
    \label{eq.feat_nor}
\end{equation}
Similarly, the target features $\boldsymbol{F}{_i^t}$ are processed to form $\tilde{\boldsymbol{F}}{^t_i} \in \mathbb{R}^{h\times w\times 1\times c}$. Subsequently, the cross-frame feature affinity $\boldsymbol{A}{_i} \in \mathbb{R}^{h\times w\times d\times d}$ in stage $i$ is calculated as follows:
\begin{equation}
    \boldsymbol{A}{_i} = 
    \operatorname{Reshape}(\tilde{\boldsymbol{F}}{^t_i}\tilde{\boldsymbol{F}}{^r_i}).
\end{equation}
The affinity volumes capture correspondences and their confidence levels between features from the two input frames. Specifically, a higher affinity value indicates a stronger resemblance between a pixel in the target frame and another pixel within the selected window of the reference frame, while a lower value suggests a mismatch or lower confidence in the correspondence.
Hence, in order to determine the relative feature position displacements, \ie, feature flow $\boldsymbol{S}{^r_i} \in \mathbb{R}^{h\times w\times 2}$, a straightforward way to localize the matched features is taking the position arguments of the maxima as follows:
\begin{equation}
    \boldsymbol{S}{^r_i} = \underset{\boldsymbol{p} \in \mathcal{W}}{\operatorname{arg} \max} \  \boldsymbol{A}{_i}(:,\boldsymbol{p}),
\end{equation}
where $\mathcal{W} = \{\boldsymbol{p}=[j,k]^\top \ | \ j,k \in [0,d]\cap\mathbb{Z}\}$ represents the set of pixels in the specified window partition.
However, the $\operatorname{argmax}$ function is non-differentiable and generates discrete outputs, which prevents the network from backpropagation and introduces quantization errors. 
To address this issue, we draw inspiration from the smooth approximation proposed in \cite{gradient2010chapelle}, and introduce a 2D $\operatorname{soft\ argmax}$ as a substitute for the original $\operatorname{argmax}$ function:
\begin{equation}
    \boldsymbol{S}{^r_i} = \sum_{\boldsymbol{p}\in \mathcal{W}} \boldsymbol{p} \frac{\exp(\boldsymbol{A}{_i}(:,\boldsymbol{p}))}{\sum_{\boldsymbol{p}}\exp(\boldsymbol{A}{_i}(:,\boldsymbol{p}))}.
\end{equation}
In this way, the position with the maximum likelihood is calculated using a probability-weighted sum of the position enumerations $\boldsymbol{p}$, where the probabilities are normalized through the softmax of the affinity values. This 2D $\operatorname{soft\ argmax}$ approach enhances feature matching with sub-pixel accuracy, facilitating the flow of gradients from pose estimation back through the point coordinates.

Another crucial piece of information conveyed by $\boldsymbol{A}{_i}$ is the confidence level $\boldsymbol{C}{_i} \in \mathbb{R}^{h\times w\times 1}$, which indicates the quality of the calculated feature flow. 
We argue that $\boldsymbol{C}{_i}$ depends on two factors: 
\begin{itemize}
    \item \textbf{Magnitude of affinity values}. If all the affinity values are relatively small, it suggests a lack of strong feature correspondences within the specified window. For example, if a moving object occupies the entire window and occludes the original matched pixel, this can result in smaller affinity values in the entire window. 
    \item \textbf{Distribution of affinity values}. If the largest affinity values are closely clustered, it suggests the presence of texture-less areas or keypoints that are difficult to discriminate. 
\end{itemize}

To avoid these aforementioned issues and lower their impact on matched correspondences, we formulate the feature matching confidence level $\boldsymbol{C}{_i}$ as follows:
\begin{equation}
    \boldsymbol{C}{_i} = \max_{\boldsymbol{p}\in\mathcal{W}}\boldsymbol{A}{_i}(:,\boldsymbol{p})\cdot
    \max_{\boldsymbol{p}\in\mathcal{W}}
    \frac{\exp(\boldsymbol{A}{_i}(:,\boldsymbol{p}))}{\sum_{\boldsymbol{p}}\exp(\boldsymbol{A}{_i}(:,\boldsymbol{p}))},
\end{equation}
where $\boldsymbol{C}{_i}$ tends to approach 1 only when there is a unique large affinity value within the given window, indicating high confidence in the feature correspondence. This formulation assists in assessing the reliability of feature matches by considering both the magnitude and the distribution of affinity values across spatial dimensions.

\subsection{Positional Clue Aggregator}
To ensure robust and effective camera pose estimation, it is essential to incorporate two primary positional clues. The first involves 2D feature flow and its corresponding pixel coordinates, which reflect the pixel-wise geometrical constraints with respect to cross-frame correlations.
To fully exploit spatial information and positional clues, we propose a positional clue aggregator, which incorporates these elements into a compact, homogeneous positional embedding space. In the $i$-th stage, the absolute feature position $\boldsymbol{S}{^a_i} \in \mathbb{R}^{h\times w\times 2}$ can be straightforwardly obtained as follows: 
\begin{equation}
    \boldsymbol{S}{^a_i} = \operatorname{Concat}(\operatorname{Meshgrid}(h,w)),
\end{equation}
where the $\operatorname{Meshgrid}$ function generates 2D grid coordinates using matrix indexing. Meanwhile, the dense point cloud $\boldsymbol{P}{^c_i} \in \mathbb{R}^{h\times w \times 3}$ in the $i$-th stage can be easily obtained from the perspective camera model.
For an arbitrary 3D point $\boldsymbol{p}{^c}=[x^c,y^c,z^c]^\top$ in camera coordinates, its corresponding homogeneous pixel coordinates $\boldsymbol{\tilde{p}}=[u,v,1]^\top$ satisfy the following relationship: 
\begin{equation}
    \boldsymbol{p}{^c} = z^c\boldsymbol{K}^{-1}\boldsymbol{\tilde{p}},
\end{equation}
where $\boldsymbol{K}$ represents the camera intrinsic matrix, and $z^c$ can be approximated using predictions from DepthNet. Thus, $\boldsymbol{P}{^c_i}$ can be generated by iterating through pixel coordinates. 

Having obtained the feature flow $\boldsymbol{S}{^r_i}$, absolute feature position $\boldsymbol{S}{^a_i}$, their corresponding confidence $\boldsymbol{C}{_i}$, and the downsampled dense point cloud $\boldsymbol{P}{^c_i}$, we proceed to encode them into a homogeneous position embedding space $\boldsymbol{F}^p_i$. First, we normalize $\boldsymbol{S}{^r_i}$, $\boldsymbol{S}{^a_i}$ and $\boldsymbol{P}{^c_i}$ into the range $[0, 1]$ using linear mapping, facilitating a uniform feature representation across different scales. Subsequently, these three positional priors are integrated into  positional embeddings $\boldsymbol{F}^p_i$ as follows:
\begin{equation}	
    \boldsymbol{F}^p_i = \boldsymbol{C}{_i}
    (
    f(\boldsymbol{\tilde{S}}{^r_i}, \mathbf{\Theta}{^s_i}) + 
    f(\boldsymbol{\tilde{S}}{^a_i}, \mathbf{\Theta}{^s_i}))
    + f(\boldsymbol{\tilde{P}}{_i}, \mathbf{\Theta}{^p_i}),
    \label{eq.pos_aggr}
\end{equation}
where $f(\cdot,\mathbf{\Theta}{_i})$ represents two consecutive convolutional layers with learnable parameters $\mathbf{\Theta}{_i}$ that map a 2D or 3D position vector into a higher embedding dimension. Notably, $\mathbf{\Theta}{^s_i}$ is shared between $\boldsymbol{S}{^r_i}$ and $\boldsymbol{S}{^a_i}$ to maintain positional encoding consistency. This approach effectively integrates spatial context from feature flow and 3D point cloud representations, enhancing the accuracy and robustness of camera pose estimation by leveraging comprehensive positional clues.

\begin{table*}[!t]
	\begin{center}
		\settablefont
		\caption{Quantitative comparison on the KITTI Eigen benchmark \cite{geiger2013vision}. In the Train column, \textbf{S} denotes training with synchronized stereo image pairs, \textbf{M} denotes training with monocular sequences, and \textbf{MS} denotes training with both monocular sequences and stereo pairs. The best results are shown in bold type.}
		\label{tb.kitti}
		\begin{tabular}{lcc|cccc|ccc}
			\toprule
			Method & training & Resolution & Abs Rel $\downarrow$ & Sq Rel $\downarrow$ & RMSE $\downarrow$ & RMSE log $\downarrow$ & $\delta <1.25$ $\uparrow$&  $\delta <1.25^2$ $\uparrow$ & $\delta <1.25^3$ $\uparrow$ \\
			\hline
			SC-DepthV3 \cite{sun2023sc}		& M & 256 $\times$ 832	& 0.118 & 0.756 & 4.709 & 0.188 & 0.864 & 0.960 & 0.984 \\
			Monodepth2 \cite{digging2019godard}		& MS & 192 $\times$ 640	& 0.106 & 0.818 & 4.750 & 0.196 & 0.874 & 0.957 & 0.979 \\
			HR-Depth \cite{hrdepth2021lyu}			& MS & 192 $\times$ 640	& 0.107 & 0.785 & 4.612 & 0.185 & 0.887 & 0.962 & 0.982 \\
			
			PackNet-SfM \cite{3d2020guizilini}		& M & 192 $\times$ 640	& 0.111 & 0.785 & 4.601 & 0.189 & 0.878 & 0.960 & 0.982 \\
			
			DIFFNet \cite{selfsupervised2021zhou}	& M & 192 $\times$ 640	& 0.102 & 0.764 & 4.483 & 0.180 & 0.890 & 0.964 & 0.983 \\
			MonoViT-tiny \cite{monovit2022zhao}		& M & 192 $\times$ 640	& 0.102 & 0.733 & 4.459 & 0.177 & 0.895 & 0.965 & 0.984 \\
			Swin-Depth \cite{swindepth2023shim}		& M & 192 $\times$ 640	& 0.106 & 0.739 & 4.510 & 0.182 & 0.890 & 0.964 & 0.984 \\
			Lite-Mono \cite{litemono2023zhang}		& M & 192 $\times$ 640	& 0.107 & 0.765 & 4.561 & 0.183 & 0.886 & 0.963 & 0.983 \\
			Lite-Mono-8M \cite{litemono2023zhang}	& M & 192 $\times$ 640	& 0.101 & 0.729 & 4.454 & 0.178 & 0.897 & 0.965 & 0.983 \\
			Dynamo-Depth \cite{dynamodepth2023sun}	& M & 192 $\times$ 640	& 0.112 & 0.758 & 4.505 & 0.183 & 0.873 & 0.959 & 0.984 \\
			ManyDepth \cite{temporal2021watson}		& M & 192 $\times$ 640	& 0.098 & 0.770 & 4.459 & 0.176 & 0.900 & 0.965 & 0.983 \\
			DynamicDepth \cite{disentangling2022feng} 					& M & 192 $\times$ 640  & 0.096 & 0.720 & 4.458 & 0.175 & 0.897 & 0.964 & 0.984 \\
			TriDepth \cite{selfsupervised2023chen}	& M & 192 $\times$ 640	& 0.093 & 0.665 & 4.272 & 0.172 & 0.907 & 0.967 & 0.984 \\
			SQLdepth \cite{wang2024sqldepth}  		& M & 192 $\times$ 640	& 0.091 & 0.713 & 4.204 & 0.169 & 0.914 & 0.968 & 0.984 \\
			
			\rowcolor{gray!20}\textbf{SCIPaD (Ours)}  & M & 192 $\times$ 640  & \textbf{0.090} &\textbf{0.650} &\textbf{4.056} &\textbf{0.166} &\textbf{0.918} &\textbf{0.970} &\textbf{0.985} \\

			\hline
			Monodepth2\cite{digging2019godard}		& MS & 320 $\times$ 1024 & 0.106 & 0.806 & 4.630 & 0.193 & 0.876 & 0.958 & 0.980 \\
			HR-Depth\cite{hrdepth2021lyu}			& MS & 320 $\times$ 1024 & 0.101 & 0.716 & 4.395 & 0.179 & 0.899 & 0.966 & 0.983 \\
			DIFFNet\cite{selfsupervised2021zhou}	& M  & 320 $\times$ 1024 & 0.097 & 0.722 & 4.435 & 0.174 & 0.907 & 0.967 & 0.984 \\
			MonoViT-tiny\cite{monovit2022zhao}		& M  & 320 $\times$ 1024 & 0.096 & 0.714 & 4.292 & 0.172 & 0.908 & 0.968 & 0.984 \\
			Lite-Mono-8M \cite{litemono2023zhang}	& M  & 320 $\times$ 1024 & 0.097 & 0.710 & 4.309 & 0.174 & 0.905 & 0.967 & 0.984 \\
			ManyDepth \cite{temporal2021watson}		& M  & 320 $\times$ 1024 & 0.087 & 0.685 & 4.142 & 0.167 & 0.920 & 0.968 & 0.983 \\
			SQLdepth\cite{wang2024sqldepth}			& M  & 320 $\times$ 1024 & 0.087 & 0.659 & 4.096 & 0.165 & 0.920 & 0.970 & 0.984 \\
			\hline
			\rowcolor{gray!20}\textbf{SCIPaD (Ours)}  & M & 320 $\times$ 1024  & \textbf{0.086} &\textbf{0.636} &\textbf{4.006} &\textbf{0.165} &\textbf{0.922} &\textbf{0.968} &\textbf{0.984} \\

			\bottomrule
		\end{tabular}
	\end{center}
\end{table*}

\begin{table*}[!t]
	\begin{center}
		\settablefont
		\caption{Quantitatve comparison using the improved KITTI ground truth provided in  \cite{uhrig2017sparsity}.}
		\label{tb.kitti_imp}
		\begin{tabular}{lcc|cccc|ccc}
			\toprule
			Method & training & Resolution & Abs Rel $\downarrow$ & Sq Rel $\downarrow$ & RMSE $\downarrow$ & RMSE log $\downarrow$ & $\delta <1.25$ $\uparrow$&  $\delta <1.25^2$ $\uparrow$ & $\delta <1.25^3$ $\uparrow$ \\
			\hline
			Monodepth2 \cite{digging2019godard}			& MS& 192 $\times$ 640	& 0.080 & 0.466 & 3.681 & 0.127 & 0.926 & 0.985 & 0.995 \\
			PackNet-SfM \cite{3d2020guizilini}			& M & 192 $\times$ 640	& 0.078 & 0.420 & 3.485 & 0.121 & 0.931 & 0.986 & 0.996 \\
			ManyDepth \cite{temporal2021watson}			& M & 192 $\times$ 640	& 0.070 & 0.399 & 3.455 & 0.113 & 0.941 & 0.989 & 0.997 \\
			DynamicDepth \cite{disentangling2022feng}	& M & 192 $\times$ 640	& 0.068 & 0.362 & 3.454 & 0.111 & 0.943 & 0.991 & 0.998 \\
			TriDepth \cite{selfsupervised2023chen}		& M & 192 $\times$ 640	& 0.068 & 0.359 & 3.341 & 0.110 & 0.944 & 0.989 & 0.997 \\
			SQLdepth \cite{wang2024sqldepth}  			& M & 192 $\times$ 640	& 0.061 & 0.317 & 3.055 & 0.100 & 0.957 & 0.992 & 0.997 \\
			
			\rowcolor{gray!20}\textbf{SCIPaD (Ours)}  		& M & 192 $\times$ 640  & \textbf{0.059} &\textbf{0.287} &\textbf{2.871} &\textbf{0.095} &\textbf{0.964} &\textbf{0.993} &\textbf{0.998} \\
			\bottomrule
		\end{tabular}
	\end{center}
\end{table*}

\subsection{Hierarchical Positional Embedding Injector}
It has been demonstrated that multi-scale feature aggregation across different modalities improves the capacity of deep neural networks in various computer vision tasks \cite{yu2018deep}.
However, as a task heavily reliant on geometric properties, camera pose estimation requires not only substantial semantic information but also accurate geometric cues. It is crucial to achieve a sensible balance between these features across different scales. This is due to the fact that the shallower layers of the network tend to contain less semantic details but richer geometric representations, whereas the deeper layers excel in capturing refined semantic abstractions but may deteriorate the meaningful spatial clues due to downsampling operations.

In this work, our proposed hierarchical positional embedding injector aims to effectively integrate low-level positional embeddings $\mathcal{F}^p$ into high-level semantic features $\mathcal{F}^s$ across different scales. Unlike previous works, such as \cite{wang2024sqldepth, sun2023sc}, which utilize the deepest features for camera pose decoding, we hierarchically aggregate fused semantic and positional features at multiple resolutions to preserve both high-level semantic and low-level positional information. For the features $\boldsymbol{F}{_i^s} \in\mathcal{F}^s $ and $\boldsymbol{F}{_i^p} \in \mathcal{F}^p$ from the $i$-th stage, we first employ a channel reduction block to transform $\boldsymbol{F}{_i^p}$ into compact embeddings. Subsequently, the compressed positional embeddings are integrated into the semantic features $\boldsymbol{F}{_i^s}$ with a learnable gate $\gamma_i$, which automatically modulates the importance of semantic and spatial information. 

The motivation for introducing the gating mechanism lies in leveraging the strengths of different network layers: the shallower layers of the network encode more precise positional embeddings, while the deeper layers preserve richer semantic information. In contrast to prior arts \cite{yu2018deep} which indiscriminately fuse the cross-modal information, our approach ensures the network adaptively focuses on semantic and positional information with different scales. Afterwards, the selectively fused features are combined with those from the preceding layer, yielding spatial-semantic co-attentive feature representations. These operations can be written as follows:
\begin{equation}
    \begin{aligned}
        \boldsymbol{F}_{i} = 
        \begin{cases}
            f_i(\gamma_i f_c(\boldsymbol{F}^{p}_{i}) + 
            (1-\gamma_i)\boldsymbol{F}^{s}_{i} + 
            \boldsymbol{F}_{i-1}, \boldsymbol{\Theta}_i),
            &\mbox{if }1\le i< k \\
            \gamma_i f_c(\boldsymbol{F}^{p}_{i}) + 
            (1-\gamma_i)\boldsymbol{F}^{s}_{i} + 
            \boldsymbol{F}_{i-1}, 
            &\mbox{if }i = k
        \end{cases}
    \end{aligned}
\label{eq.hpei}
\end{equation}
where $k$ denotes the number of stages in the backbone, $\gamma_i$ regulates the importance between semantic features $\boldsymbol{F}^{s}_{i}$ and positional features $\boldsymbol{F}^{p}_{i}$, $f_i(\cdot,\boldsymbol{\Theta_i})$ represents a combination of convolutional, rectified linear unit (ReLU), and downsampling layers with learnable parameters $\boldsymbol{\Theta}_i$, and $f_c(\cdot)$ denotes the channel reduction function performed by a convolutional layer with a kernel size of 1$\times$1. $\boldsymbol{F}_{0}$ is initialized as a zero matrix with the same shape of $\boldsymbol{F}^{s}_{1}$, and the final aggregated output $\boldsymbol{F}_{4}$ is subsequently passed through a multilayer perceptron (MLP) for camera pose decoding.

\begin{figure*}
	\centering
	\includegraphics[width=0.99\linewidth]{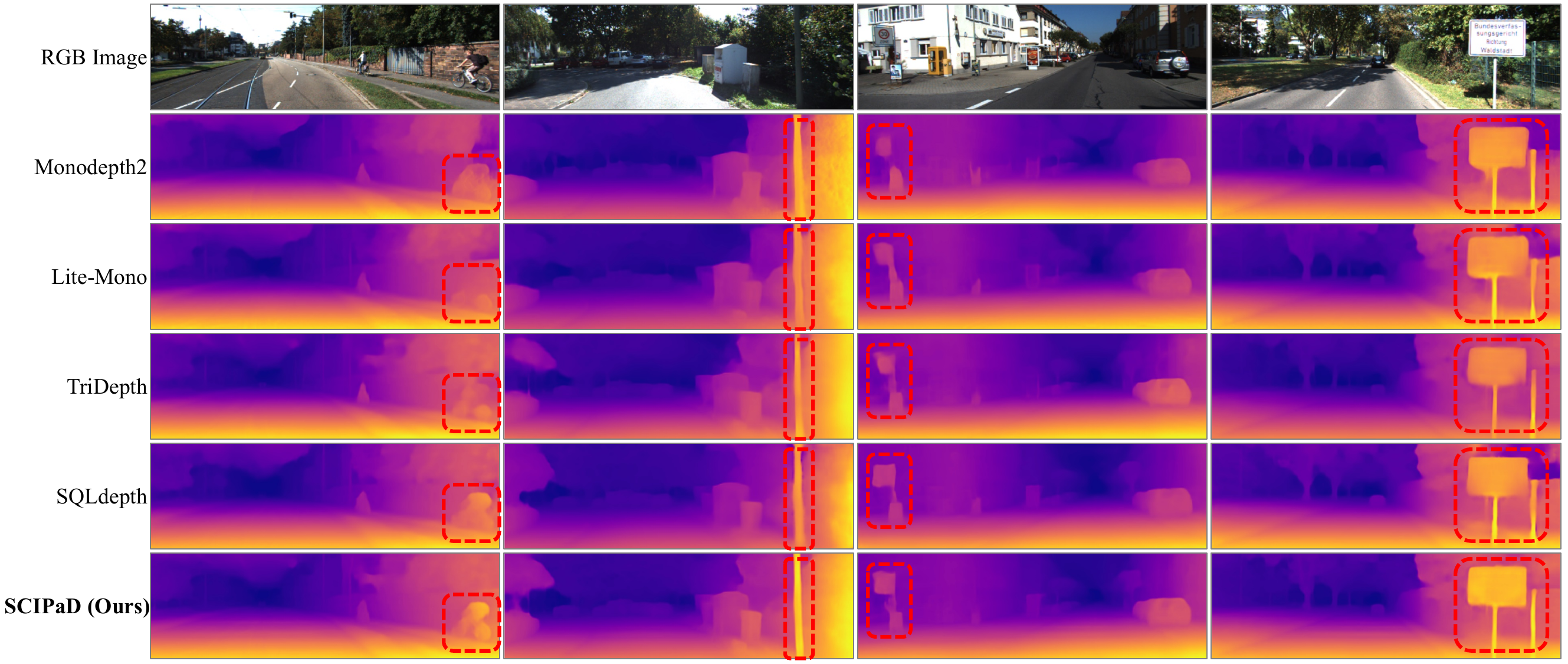}
	\caption{Qualitative experimental results on the KITTI Eigen benchmark. The regions highlighted in the red boxes illustrate that our method produces locally consistent depth maps with enhanced details.}
	\label{fig.kitti}
\end{figure*}

\section{Experiments}
\label{sec.exp}
\subsection{Datasets, Evaluation Metrics, and Implementation Details}
Our proposed method is evaluated on three public datasets, including the KITTI Raw dataset \cite{are2012geiger}, the KITTI Odometry dataset \cite{geiger2013vision}, and the Make3D dataset \cite{make3d2009saxena}. Specifically, we evaluate the depth estimation performance on the KITTI Raw dataset, and evaluate the camera pose estimation results on the KITTI Odometry dataset. Moreover, we also assess the generalizability of our model on the Make3D dataset using the weights pretrained on the KITTI Raw dataset.

\begin{itemize}
    \item \textbf{KITTI Raw} \cite{are2012geiger}: This dataset consists of driving videos recorded in urban environments, and it is widely used in self-supervised monocular depth estimation research. Following previous works \cite{sun2023sc, wang2024sqldepth, unsupervised2017zhou, selfsupervised2019chen}, we adopt the Eigen split, which uses 39,810 images for training, 4,424 images for evaluation, and 697 images for testing. Depth ranges are capped at 80 m, and all images are resized to the resolution of $192 \times 640$ pixels or $320 \times 1024$ pixels for network training. 
    \item \textbf{KITTI Odometry} \cite{geiger2013vision}: This dataset contains well-rectified stereo images with ground-truth trajectories in 22 driving scenarios. Following previous work \cite{digging2019godard}, we use Seqs. 00-08 for model training and test our method on Seqs. 09-10. The absolute trajectory error is calculated by averaging over all overlapping five-frame snippets in the test sequences, following the approach proposed in \cite{digging2019godard}.
    \item \textbf{Make3D} \cite{make3d2009saxena}: We evaluate the generalizability of the proposed method on the Make3D dataset, which contains 134 test images of outdoor scenes. 
    The proposed model, initially trained on the KITTI Raw dataset, is directly applied to these test images for evaluation.
    
\end{itemize}

% see undeepvo, towards better generalization
% kitti odom
\begin{table}[!t]
	\begin{center}
		\settablefont
		\caption{Visual odometry results on the KITTI Odometry dataset \cite{geiger2013vision}. }
		\label{tb.kitti_odom}
		
		\begin{tabular}{l|ccc|ccc}
			\toprule
			\multirow{2}{*}{Method}      & \multicolumn{3}{c|}{Seq. 09}        & \multicolumn{3}{c}{Seq. 10}       \\
			& $e_t$ (\%) & $e_r$ (\%)  & ATE (m) & $e_t$ (\%) & $e_r$ (\%)  & ATE (m) \\
			\hline
			SfMLearner \cite{unsupervised2017zhou} 	&19.15	&6.82 &77.79  &40.40 &17.69 &67.34\\
			GeoNet \cite{geonet2018yin}      		&28.72  &9.80 &158.45 &23.90 &9.00  &43.04\\
			DeepMatchVO \cite{shen2019beyond}       &9.91 	&3.80 &27.08  &12.18 &5.90  &24.44\\
			Monodepth2 \cite{digging2019godard}		&36.70  &16.36 &99.14  &49.71 &25.08  &86.94\\
			SC-Depth \cite{unsupervised2021bian}	&12.16	&4.01 &58.79  &12.23 &6.20  &16.42\\
			\hline
			\rowcolor{gray!20}
			\textbf{SCIPaD (Ours)}  				&\textbf{7.43}	&\textbf{2.46} &\textbf{26.15}  &\textbf{9.82} &\textbf{3.87}  &\textbf{15.51}\\
			\bottomrule
		\end{tabular}
	\end{center}
\end{table}

Adhering to the experiments presented in the previous works \cite{unsupervised2017zhou, digging2019godard}, we quantify the model's performance using the mean absolute relative error (Abs Rel), the mean squared relative error (Sq Rel), the root mean squared error (RMSE), the root mean squared log error (RMSE log), and the accuracy under threshold ($\delta_i < 1.25^i$, $i = 1, 2, 3$). Detailed definitions of these metrics can be found in \cite{depth2014eigen}. For visual odometry performance evaluation, we follow the standard evaluation metrics introduced in \cite{geiger2013vision}, including the average translational error $e_t$ (\%), the average rotational error $e_r$ (\%), and the absolute trajectory error ATE (m) \cite{sturm2012benchmark}.

The proposed method is implemented in PyTorch and trained on an NVIDIA RTX 4090 GPU. We adopt the TriDepth \cite{selfsupervised2023chen} framework as our baseline. Following \cite{digging2019godard}, we utilize a snippet of three sequential video frames as a training sample. During training, images are augmented with random color jitter and horizontal flips. We employ the Adam optimizer \cite{kingma2014adam} and conduct the training over 30 epochs, starting with a learning rate of $10^{-4}$, which we reduce by a factor of 10 for the final 5 epochs. In line with practices from \cite{digging2019godard, wang2024sqldepth, temporal2021watson}, we initialize the encoder of our network using pretrained weights from ImageNet \cite{imagenet2009deng}.

\begin{table*}[!t]
	\begin{center}
		\settablefont
		\caption{Comparisons of existing models with and without our proposed PoseNet embedded on the KITTI Eigen benchmark. Manual replications with released code are indicated by the symbol $^\dagger$.}
		\label{tb.abl_posenet}
		\begin{tabular}{lcc|cccc|ccc}
			\toprule
			Method  & with Ours & Resolution (Pixels)  & Abs Rel $\downarrow$ & Sq Rel $\downarrow$ & RMSE $\downarrow$ & RMSE log $\downarrow$ & $\delta <1.25$ $\uparrow$&  $\delta <1.25^2$ $\uparrow$ & $\delta <1.25^3$ $\uparrow$ \\
			\hline
			\multirow{2}*{SC-DepthV3$^\dagger$ \cite{sun2023sc}}	
			&			& 256 $\times$ 832	&0.119 &0.756 &4.699 &0.188 &0.863 &0.960 &0.984 \\
			&\checkmark & 256 $\times$ 832	&0.113 &0.742 &4.603 &0.185 &0.870 &0.962 &0.984 \\
			\hline
			\multirow{2}*{Monodepth2$^\dagger$ \cite{digging2019godard}}	
			&			& 192 $\times$ 640	&0.116 &0.923 &4.856 &0.193	&0.878 &0.959 &0.981 \\
			&\checkmark & 192 $\times$ 640	&0.115 &0.902 &4.827 &0.192 &0.877 &0.959 &0.981 \\
			\hline
			\multirow{2}*{ManyDepth$^\dagger$ \cite{temporal2021watson}}	
			&			& 192 $\times$ 640	&0.101  &0.800  &4.583  &0.182  &0.894  &0.962  &0.982  \\ 
			&\checkmark & 192 $\times$ 640	&0.098  &0.758  &4.430  &0.177  &0.899  &0.964  &0.983  \\
			\hline
			\multirow{2}*{TriDepth$^\dagger$ \cite{selfsupervised2023chen}}	
			&			& 192 $\times$ 640	&0.095	&0.718	&4.408	&0.176	&0.896	&0.964	&0.983 \\				
			&\checkmark & 192 $\times$ 640	&0.094	&0.699	&4.325	&0.167	&0.903	&0.969	&0.984	\\				
			\bottomrule
		\end{tabular}
	\end{center}
\end{table*}

\begin{table*}[!t]
	\begin{center}
		\settablefont
		\caption{Ablation studies of SCIPaD design on the KITTI Raw and KITTI Odometry datasets.}
		\label{tb.abl_inner}
		\begin{tabular}{cc|cccc|ccc|ccc}
			\toprule
			\multicolumn{2}{c|}{\multirow{3}{*}{Configuration}} 	& \multicolumn{4}{c|}{\multirow{2}{*}{KITTI Raw Dataset}} & \multicolumn{6}{c}{KITTI Odometry Dataset}                                         
			\\ \cline{7-12} 
			&&&&&	& \multicolumn{3}{c|}{Seq. 09}	& \multicolumn{3}{c}{Seq. 10}
			\\ \cline{3-12} 
			&& Abs Rel  $\downarrow$ & $\delta_1$ $\uparrow$&  $\delta_2$ $\uparrow$ & $\delta_3$ $\uparrow$ 
			& $e_t$ (\%) & $e_r$ (\%)  & ATE (m) & $e_t$ (\%) & $e_r$ (\%)  & ATE (m)   \\ \hline \rowcolor{gray!20}
			
			\multicolumn{2}{c|}{Full Implementation}							&\textbf{0.090}	&0.918	&\textbf{0.970}	&\textbf{0.985}	&7.43	&2.46	&26.15	&\textbf{9.82}	&\textbf{3.87}	&\textbf{15.51}  \\ \hline
			\multicolumn{1}{c|}{\multirow{3}{*}{CAFFE}} &Unfrozen Backbone		&0.090	&\textbf{0.919}	&0.969	&0.985	&\textbf{7.37}	&\textbf{2.10}	&\textbf{24.84}	&12.53	&4.84	&19.26  \\
			\multicolumn{1}{c|}{} &w/o Confidence Reweighting 					&0.092	&0.912	&0.967	&0.984	&10.92	&3.03	&45.61	&12.49	&6.31	&25.83  \\
			\multicolumn{1}{c|}{} &w/o Feature Normalization 					&0.091	&0.917	&0.968	&0.984	&9.84	&2.89	&41.84	&12.97	&6.94	&27.42  \\ \hline
			\multicolumn{1}{c|}{\multirow{4}{*}{PCA}} &w/o Feature Flow 		&0.095	&0.907	&0.965	&0.984	&15.38	&5.26	&60.05	&23.39	&9.55	&43.29  \\
			\multicolumn{1}{c|}{} &w/o Absolute Feature Position 				&0.091	&0.914	&0.968	&0.984	&10.01	&2.98	&41.84	&12.97	&6.94	&27.42  \\
			\multicolumn{1}{c|}{} &w/o Depth Predictions 						&0.092	&0.913	&0.967	&0.984	&10.52	&3.12	&46.51	&13.55	&6.72	&28.81  \\
			\multicolumn{1}{c|}{} &w/ Shared Embedding Layer 					&0.094	&0.903	&0.965	&0.984	&16.84	&3.51	&55.36	&18.45	&8.36	&39.82  \\ \hline
			\multicolumn{1}{c|}{\multirow{4}{*}{HPEI}} &w/o Gating Mechanism 	&0.092	&0.913	&0.967	&0.984	&10.42	&3.09	&46.85	&13.75	&6.54	&26.88  \\
			\multicolumn{1}{c|}{} &Decode $\mathcal{F}^s$ Only 					&0.093	&0.907	&0.967	&0.984	&15.24	&3.49	&52.23	&15.42	&7.23	&30.14  \\
			\multicolumn{1}{c|}{} &Decode $\mathcal{F}^p$ Only 					&0.096	&0.897	&0.964	&0.983	&12.14	&4.52	&52.83	&12.41	&6.33	&35.64  \\ \bottomrule
		\end{tabular}
	\end{center}
\end{table*}

\subsection{Comparison with State-of-The-Arts}
\textbf{Depth Estimation Results.}
As presented in Table \ref{tb.kitti}, our proposed SCIPaD achieves SoTA performance on both resolutions on the KITTI Raw dataset. It can be observed that SCIPaD demonstrates superior performance compared to all existing self-supervised methods, achieving an 8.8\% reduction in Sq Rel, and a 0.4\% improvement in $\delta_1$ compared to SQLdepth \cite{wang2024sqldepth}, a previous SoTA approach. Moreover, SCIPaD significantly outperforms its counterparts trained with additional stereo image pairs such as Monodepth2 \cite{digging2019godard} and HR-Depth \cite{hrdepth2021lyu}, achieving an average error reduction of 15.5\% in Abs Rel and an average performance gain of 1.1\% in $\delta_1$. 

Fig. \ref{fig.kitti} shows the qualitative experimental results on the KITTI Eigen benchmark. Compared with previous SoTA methods, which often produce blurry edges in the foreground, our method generates more distinctive boundaries (\eg, the first and third columns) and maintains better depth consistency in continuous regions (\eg, the second and fourth columns). This superior performance is primarily due to our method's ability to determine accurate and robust camera poses, which in turn generate more refined self-supervisory cues for photometric reconstruction.

Due to the limited quality of the original ground-truth data in KITTI, we additionally present evaluation results using the improved KITTI ground truth \cite{uhrig2017sparsity} in Table \ref{tb.kitti_imp}. Even when compared with the previous SoTA method SQLdepth \cite{wang2024sqldepth}, SCIPaD continues to demonstrate superior performance across all metrics. Notably, it achieves a 3.3\% reduction in Abs Rel and a 9.5\% reduction in Sq Rel at a resolution of 640$\times$192 pixels.

\textbf{Camera Pose Estimation Results.}
We use the KITTI Odometry dataset \cite{geiger2013vision} to evaluate the camera pose estimation performance of our proposed SCIPaD. 
Models trained on monocular videos often struggle to recover absolute depth metrics due to scale ambiguity. To address this issue, we align the scale of their predicted results with the ground truth using 7-DoF optimization.
As presented in Table \ref{tb.kitti_odom}, SCIPaD significantly outperforms other monocular visual odometry methods. Compared to the previous SoTA frameworks, our proposed method achieves a reduction of 22.2\% in $e_t$, 34.8\% in $e_r$, and 4.5\% in ATE, respectively. 

\begin{figure}
	\centering
	\includegraphics[width=0.99\linewidth]{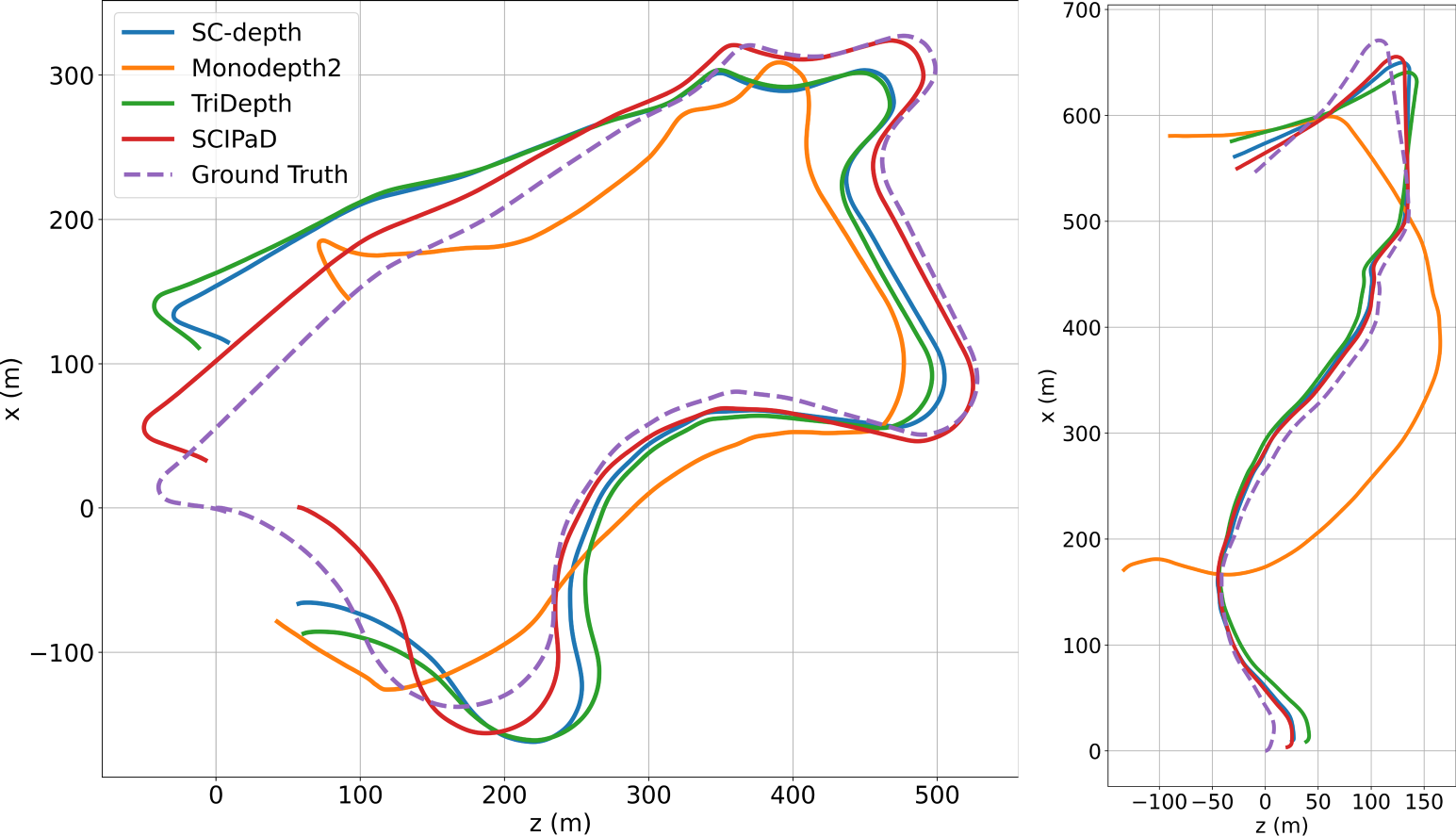}
	\caption{Comparison of the estimated trajectories using Seqs. 09 and 10 on the KITTI Odometry dataset \cite{geiger2013vision}. All predictions are rescaled to align with the ground truth for a fair comparison. }
	\label{fig.vis_odom}
\end{figure}

We also provide a qualitative comparison of the trajectories produced by different methods. As shown in Fig. \ref{fig.vis_odom}, we evaluate monocular visual odometry methods on Seq. 09 (left) and Seq. 10 (right) of the KITTI Odometry dataset, and SCIPaD exhibits superior results, with minimal drift among all SoTA methods. This demonstrates the high performance and robustness of our method in ego-motion recovery and long-term trajectory estimation.

\begin{figure*}
	\centering
	\includegraphics[width=0.99\linewidth]{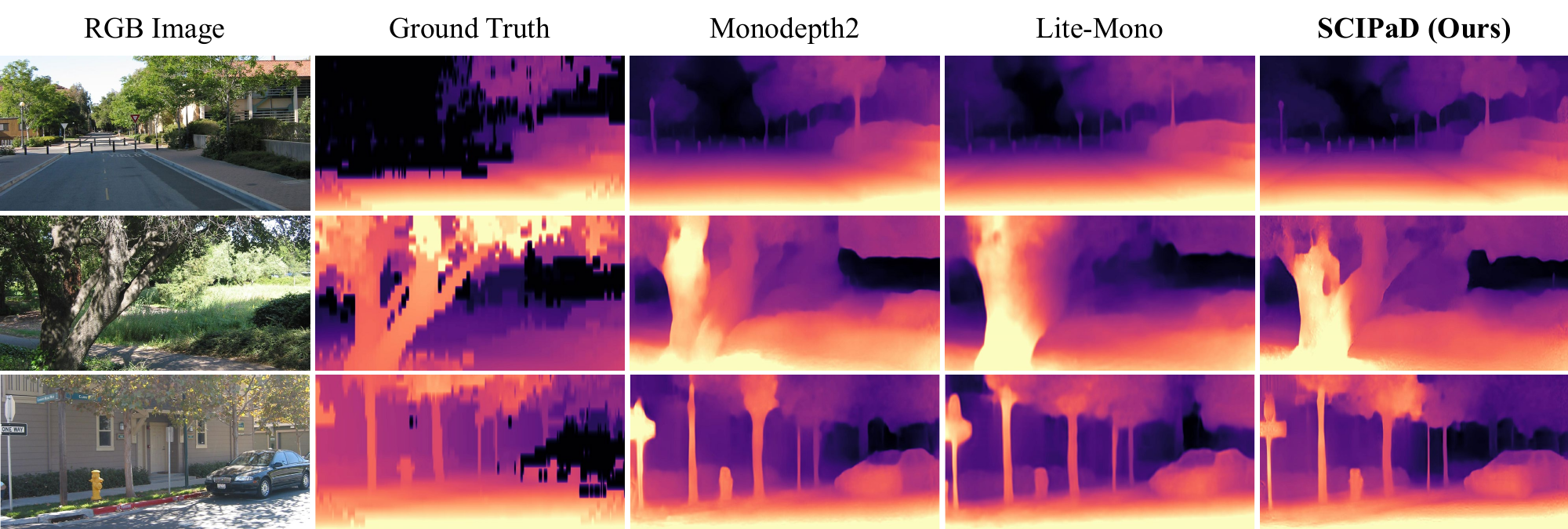}
	\caption{Qualitative zero-shot results on the Make3D dataset \cite{make3d2009saxena}.}
	\label{fig.make3d}
\end{figure*}

\subsection{Ablation Study}
As shown in Table \ref{tb.abl_posenet}, we incorporate the PoseNet in SCIPaD into existing open-source methods and demonstrate that our method significantly enhances their depth estimation performances. Remarkably, this integration results in a substantial performance boost across all metrics for the original models.

Furthermore, we investigate the rationality and efficacy of our proposed SCIPaD. As illustrated in Table \ref{tb.abl_inner}, we conduct ablation experiments regarding the inner design of CAFFE, PCA, and HPEI. First, we observe that unfreezing the feature flow backbone results in minor performance gains but significantly increases the computational burden. Therefore, we opt to maintain a frozen feature flow to achieve a balance between accuracy and computational efficiency. Moreover, we notice that the removal of feature normalization in (\ref{eq.feat_nor}) and confidence reweighting in (\ref{eq.pos_aggr}) leads to reduced performance across the two datasets, confirming the necessity of these components. Second, feature flow, absolute feature position, and DepthNet predictions are three spatial clues to be aggregated. Removing these elements sequentially highlighted that feature flow is most critical for depth estimation and ego-motion recovery, while the absolute feature position contributes the least. Including DepthNet predictions enhances pose estimation significantly. Using the same embedding layer for aggregating all three spatial clues, as noted in (\ref{eq.pos_aggr}), led to a noticeable performance decline, suggesting the need for distinct processing of each clue. Third, we evaluate the efficacy of the gating mechanism in (\ref{eq.hpei}) as well as the overall contributions of semantic features $\mathcal{F}^s$ and positional embeddings $\mathcal{F}^p$. The results indicate that both components are crucial to the system’s performance, significantly impacting the depth-pose joint estimation results.

\subsection{Zero-Shot Performance Evaluation}
To further evaluate the generalizability of SCIPaD, we conduct a zero-shot test on the Make3D dataset \cite{make3d2009saxena} using the pretrained weights obtained from the KITTI dataset. Following the evaluation settings used in \cite{wang2024sqldepth}, the test images are center-cropped to a 2:1 ratio for a fair comparison. As presented in Table \ref{tb.make3d} and Fig. \ref{fig.make3d}, SCIPaD outperforms other methods in zero-shot performance, and produces finer-grained depth maps with more accurate scene details. These results demonstrate the exceptional zero-shot generalizability of our model.

\begin{table}[!t]
    \begin{center}
        \settablefont
        \caption{Quantitative zero-shot performance comparison on the Make3D dataset \cite{make3d2009saxena}.}
        \label{tb.make3d}
        \begin{tabular}{c|cccc}
            \toprule
            Method     & Abs Rel & Sq Rel & RMSE  & RMSE log \\ \hline
            Monodepth2 \cite{digging2019godard} 	& 0.321   & 3.378  & 7.252 & 0.163    \\
            HR-Depth \cite{hrdepth2021lyu}  		& 0.305   & 2.944  & 6.857 & 0.157    \\
            CADepth \cite{yan2021channel}   		& 0.319   & 3.564  & 7.152 & 0.158    \\
            DIFFNet \cite{selfsupervised2021zhou}   & 0.298   & 2.901  & 6.753 & 0.153    \\
            MonoViT \cite{monovit2022zhao}   		& 0.286   & 2.758  & 6.623 & 0.147    \\
            \hline \rowcolor{gray!20}
            \textbf{SCIPaD (Ours)}					& 0.284   & 2.712  & 6.593 & 0.147    \\
            \bottomrule
        \end{tabular}
    \end{center}
\end{table}

\section{Conclusion}
\label{sec.conclusion}
In this article, we introduced SCIPaD, a novel architecture designed for unsupervised learning of ego-motion and monocular depth estimation. SCIPaD estimates confidence-aware feature flow from a CAFFE, and aggregates spatial clues into homogeneous positional representations using a PCA. Finally, an HPEI selectively injects positional embeddings into semantic information for robust ego-motion decoding. Our proposed method achieves remarkable state-of-the-art results and improved generalizability on the KITTI Raw, KITTI Odometry, and Make3D datasets. Future work will focus on developing a lightweight version of SCIPaD and further enhancing the generalizability performance of the depth-pose joint learning framework.

\bibliographystyle{IEEEtran} 
\bibliography{ref}

\begin{thebibliography}{10}
\providecommand{\url}[1]{#1}
\csname url@samestyle\endcsname
\providecommand{\newblock}{\relax}
\providecommand{\bibinfo}[2]{#2}
\providecommand{\BIBentrySTDinterwordspacing}{\spaceskip=0pt\relax}
\providecommand{\BIBentryALTinterwordstretchfactor}{4}
\providecommand{\BIBentryALTinterwordspacing}{\spaceskip=\fontdimen2\font plus
\BIBentryALTinterwordstretchfactor\fontdimen3\font minus
  \fontdimen4\font\relax}
\providecommand{\BIBforeignlanguage}[2]{{%
\expandafter\ifx\csname l@#1\endcsname\relax
\typeout{** WARNING: IEEEtran.bst: No hyphenation pattern has been}%
\typeout{** loaded for the language `#1'. Using the pattern for}%
\typeout{** the default language instead.}%
\else
\language=\csname l@#1\endcsname
\fi
#2}}
\providecommand{\BIBdecl}{\relax}
\BIBdecl

\bibitem{li2024roadformer}
J.~Li \emph{et~al.}, ``{RoadFormer: Duplex Transformer for RGB-Normal Semantic
  Road Scene Parsing},'' \emph{IEEE Transactions on Intelligent Vehicles},
  2024, {DOI}: 10.1109/TIV.2024.3388726.

\bibitem{fan2020sne}
R.~Fan \emph{et~al.}, ``{SNE-RoadSeg: Incorporating Surface Normal Information
  into Semantic Segmentation for Accurate Freespace Detection},'' in
  \emph{Proceedings of the European Conference on Computer Vision
  (ECCV)}.\hskip 1em plus 0.5em minus 0.4em\relax Springer, 2020, pp. 340--356.

\bibitem{fan2019road}
R.~Fan and M.~Liu, ``Road damage detection based on unsupervised disparity map
  segmentation,'' \emph{IEEE Transactions on Intelligent Transportation
  Systems}, vol.~21, no.~11, pp. 4906--4911, 2020.

\bibitem{fan2021graph}
R.~Fan \emph{et~al.}, ``Graph attention layer evolves semantic segmentation for
  road pothole detection: A benchmark and algorithms,'' \emph{IEEE Transactions
  on Image Processing}, vol.~30, pp. 8144--8154, 2021.

\bibitem{feng2023freespace}
Y.~Feng \emph{et~al.}, ``{Freespace Optical Flow Modeling for Automated
  Driving},'' \emph{IEEE/ASME Transactions on Mechatronics}, vol.~29, no.~2,
  pp. 1511--1520, 2024.

\bibitem{digging2019godard}
C.~Godard \emph{et~al.}, ``{Digging into Self-Supervised Monocular Depth
  Estimation},'' in \emph{Proceedings of the IEEE/CVF International Conference
  on Computer Vision (ICCV)}, 2019, pp. 3828--3838.

\bibitem{fan2018road}
R.~Fan \emph{et~al.}, ``{Road Surface 3D Reconstruction Based on Dense Subpixel
  Disparity Map Estimation},'' \emph{IEEE Transactions on Image Processing},
  vol.~27, no.~6, pp. 3025--3035, 2018.

\bibitem{feng2023d2nt}
Y.~Feng \emph{et~al.}, ``{D2NT: A High-Performing Depth-to-Normal
  Translator},'' in \emph{IEEE International Conference on Robotics and
  Automation (ICRA)}.\hskip 1em plus 0.5em minus 0.4em\relax IEEE, 2023, pp.
  12\,360--12\,366.

\bibitem{depth2014eigen}
D.~Eigen \emph{et~al.}, ``{Depth Map Prediction from a Single Image Using a
  Multi-Scale Deep Network},'' in \emph{Advances in Neural Information
  Processing Systems (NeurIPS)}, vol.~27, 2014.

\bibitem{learning2016liu}
F.~Liu \emph{et~al.}, ``{Learning Depth from Single Monocular Images Using Deep
  Convolutional Neural Fields},'' \emph{IEEE Transactions on Pattern Analysis
  and Machine Intelligence}, vol.~38, no.~10, pp. 2024--2039, 2016.

\bibitem{huang2024online}
Z.~Huang \emph{et~al.}, ``{Online, Target-Free Lidar-Camera Extrinsic
  Calibration via Cross-Modal Mask Matching},'' \emph{IEEE Transactions on
  Intelligent Vehicles}, 2024, in press.

\bibitem{zhao2023dive}
H.~Zhao \emph{et~al.}, ``{Dive Deeper into Rectifying Homography for Stereo
  Camera Online Self-Calibration},'' \emph{2024 International Conference on
  Robotics and Automation (ICRA)}, 2024, in press.

\bibitem{wang2024sqldepth}
Y.~Wang \emph{et~al.}, ``{SQLdepth: Generalizable Self-Supervised
  Fine-Structured Monocular Depth Estimation},'' in \emph{Proceedings of the
  AAAI Conference on Artificial Intelligence (AAAI)}, vol.~38, no.~6, 2024, pp.
  5713--5721.

\bibitem{posenet2015kendall}
A.~Kendall \emph{et~al.}, ``{PoseNet: a Convolutional Network for Real-Time
  6-Dof Camera Relocalization},'' in \emph{Proceedings of the IEEE/CVF
  International Conference on Computer Vision (ICCV)}, 2015, pp. 2938--2946.

\bibitem{campos2021orb}
C.~Campos \emph{et~al.}, ``{ORB-SLAM3: an Accurate Open-Source Library for
  Visual, Visual--Inertial, and Multimap Slam},'' \emph{IEEE Transactions on
  Robotics}, vol.~37, no.~6, pp. 1874--1890, 2021.

\bibitem{qin2018vins}
T.~Qin \emph{et~al.}, ``{VINS-Mono: a Robust and Versatile Monocular
  Visual-Inertial State Estimator},'' \emph{IEEE Transactions on Robotics},
  vol.~34, no.~4, pp. 1004--1020, 2018.

\bibitem{imagenet2009deng}
J.~Deng \emph{et~al.}, ``{ImageNet: a Large-Scale Hierarchical Image
  Database},'' in \emph{Proceedings of the IEEE/CVF Conference on Computer
  Vision and Pattern Recognition (CVPR)}, 2009, pp. 248--255.

\bibitem{jia2024feature}
N.~Jia \emph{et~al.}, ``{TFGNet}: Traffic salient object detection using a
  feature deep interaction and guidance fusion,'' \emph{IEEE Transactions on
  Intelligent Transportation Systems}, vol.~25, no.~3, pp. 3020--3030, 2024.

\bibitem{wu2024s}
Z.~Wu \emph{et~al.}, ``{S$^3$M-Net: Joint Learning of Semantic Segmentation and
  Stereo Matching for Autonomous Driving},'' \emph{IEEE Transactions on
  Intelligent Vehicles}, vol.~9, no.~2, pp. 3940--3951, 2024.

\bibitem{xue2020toward}
F.~Xue \emph{et~al.}, ``{Toward Hierarchical Self-Supervised Monocular Absolute
  Depth Estimation for Autonomous Driving Applications},'' in \emph{IEEE/RSJ
  International Conference on Intelligent Robots and Systems (IROS)}.\hskip 1em
  plus 0.5em minus 0.4em\relax IEEE, 2020, pp. 2330--2337.

\bibitem{3d2020guizilini}
V.~Guizilini \emph{et~al.}, ``{3D Packing for Self-Supervised Monocular Depth
  Estimation},'' in \emph{Proceedings of the IEEE/CVF Conference on Computer
  Vision and Pattern Recognition (CVPR)}, 2020, pp. 2485--2494.

\bibitem{el2019survey}
F.~El~Jamiy and R.~Marsh, ``{Survey on Depth Perception in Head Mounted
  Displays: Distance Estimation in Virtual Reality, Augmented Reality, and
  Mixed Reality},'' \emph{IEEE Transactions on Image Processing}, vol.~13,
  no.~5, pp. 707--712, 2019.

\bibitem{li2020unsupervised}
L.~Li \emph{et~al.}, ``{Unsupervised-Learning-Based Continuous Depth and Motion
  Estimation with Monocular Endoscopy for Virtual Reality Minimally Invasive
  Surgery},'' \emph{IEEE Transactions on Robotics}, vol.~17, no.~6, pp.
  3920--3928, 2020.

\bibitem{zhang2024dcpi}
M.~Zhang \emph{et~al.}, ``{DCPI-Depth}: Explicitly infusing dense
  correspondence prior to unsupervised monocular depth estimation,''
  \emph{CoRR}, 2024.

\bibitem{estimating2018cao}
Y.~Cao \emph{et~al.}, ``{Estimating Depth from Monocular Images as
  Classification Using Deep Fully Convolutional Residual Networks},''
  \emph{IEEE Transactions on Circuits and Systems for Video Technology},
  vol.~28, no.~11, pp. 3174--3182, 2018.

\bibitem{monocular2021song}
M.~Song \emph{et~al.}, ``{Monocular Depth Estimation Using Laplacian
  Pyramid-Based Depth Residuals},'' \emph{IEEE Transactions on Circuits and
  Systems for Video Technology}, vol.~31, no.~11, pp. 4381--4393, 2021.

\bibitem{enforcing2019yin}
W.~Yin \emph{et~al.}, ``{Enforcing Geometric Constraints of Virtual Normal for
  Depth Prediction},'' in \emph{Proceedings of the IEEE/CVF International
  Conference on Computer Vision (ICCV)}, 2019, pp. 5684--5693.

\bibitem{deep2018fu}
H.~Fu \emph{et~al.}, ``{Deep Ordinal Regression Network for Monocular Depth
  Estimation},'' in \emph{Proceedings of the IEEE/CVF Conference on Computer
  Vision and Pattern Recognition (CVPR)}, 2018, pp. 2002--2011.

\bibitem{big2021lee}
J.~H. Lee \emph{et~al.}, ``{From Big to Small: Multi-Scale Local Planar
  Guidance for Monocular Depth Estimation},'' \emph{CoRR}, 2021.

\bibitem{adabins2021bhat}
S.~F. Bhat \emph{et~al.}, ``{AdaBins: Depth Estimation Using Adaptive Bins},''
  in \emph{Proceedings of the IEEE/CVF Conference on Computer Vision and
  Pattern Recognition (CVPR)}, 2021, pp. 4009--4018.

\bibitem{transformerbased2021yang}
G.~Yang \emph{et~al.}, ``{Transformer-Based Attention Networks for Continuous
  Pixel-Wise Prediction},'' in \emph{Proceedings of the IEEE/CVF International
  Conference on Computer Vision (ICCV)}, 2021, pp. 16\,269--16\,279.

\bibitem{depth2024yang}
L.~Yang \emph{et~al.}, ``{Depth Anything: Unleashing the Power of Large-Scale
  Unlabeled Data},'' in \emph{Proceedings of the IEEE/CVF Conference on
  Computer Vision and Pattern Recognition (CVPR)}, 2024.

\bibitem{birklMiDaSV3Model2023}
R.~Birkl \emph{et~al.}, ``{MiDaS v3.1 -- a Model Zoo for Robust Monocular
  Relative Depth Estimation},'' \emph{CoRR}, 2023.

\bibitem{dinov22023oquab}
M.~Oquab \emph{et~al.}, ``{DINOv2: Learning Robust Visual Features without
  Supervision},'' \emph{Transactions on Machine Learning Research}, 2024.

\bibitem{unsupervised2016garg}
R.~Garg \emph{et~al.}, ``{Unsupervised CNN for Single View Depth Estimation:
  Geometry to the Rescue},'' in \emph{Computer Vision -- ECCV 2016}, B.~Leibe,
  J.~Matas, N.~Sebe, and M.~Welling, Eds.\hskip 1em plus 0.5em minus
  0.4em\relax Springer International Publishing, 2016, pp. 740--756.

\bibitem{unsupervised2017godard}
C.~Godard \emph{et~al.}, ``{Unsupervised Monocular Depth Estimation with
  Left-Right Consistency},'' in \emph{Proceedings of the IEEE/CVF Conference on
  Computer Vision and Pattern Recognition (CVPR)}, 2017, pp. 270--279.

\bibitem{unsupervised2017zhou}
T.~Zhou \emph{et~al.}, ``{Unsupervised Learning of Depth and Ego-Motion from
  Video},'' in \emph{Proceedings of the IEEE/CVF Conference on Computer Vision
  and Pattern Recognition (CVPR)}, 2017, pp. 1851--1858.

\bibitem{geonet2018yin}
Z.~Yin and J.~Shi, ``{GeoNet: Unsupervised Learning of Dense Depth, Optical
  Flow and Camera Pose},'' in \emph{Proceedings of the IEEE/CVF Conference on
  Computer Vision and Pattern Recognition (CVPR)}, 2018, pp. 1983--1992.

\bibitem{competitive2019ranjan}
A.~Ranjan \emph{et~al.}, ``{Competitive Collaboration: Joint Unsupervised
  Learning of Depth, Camera Motion, Optical Flow and Motion Segmentation},'' in
  \emph{Proceedings of the IEEE/CVF Conference on Computer Vision and Pattern
  Recognition (CVPR)}, 2019, pp. 12\,240--12\,249.

\bibitem{selfsupervised2023chen}
X.~Chen \emph{et~al.}, ``{Self-Supervised Monocular Depth Estimation: Solving
  the Edge-Fattening Problem},'' in \emph{Proceedings of the IEEE/CVF Winter
  Conference on Applications of Computer Vision (WACV)}, 2023, pp. 5776--5786.

\bibitem{finegrained2021jung}
H.~Jung \emph{et~al.}, ``{Fine-Grained Semantics-Aware Representation
  Enhancement for Self-Supervised Monocular Depth Estimation},'' in
  \emph{Proceedings of the IEEE/CVF International Conference on Computer Vision
  (ICCV)}, 2021, pp. 12\,642--12\,652.

\bibitem{uncertainty2020poggi}
M.~Poggi \emph{et~al.}, ``{On the Uncertainty of Self-Supervised Monocular
  Depth Estimation},'' in \emph{Proceedings of the IEEE/CVF Conference on
  Computer Vision and Pattern Recognition (CVPR)}, 2020, pp. 3227--3237.

\bibitem{d3vo2020yang}
N.~Yang \emph{et~al.}, ``{D3VO: Deep Depth, Deep Pose and Deep Uncertainty for
  Monocular Visual Odometry},'' in \emph{Proceedings of the IEEE/CVF Conference
  on Computer Vision and Pattern Recognition (CVPR)}, 2020, pp. 1281--1292.

\bibitem{temporal2021watson}
J.~Watson \emph{et~al.}, ``{The Temporal Opportunist: Self-Supervised
  Multi-Frame Monocular Depth},'' in \emph{Proceedings of the IEEE/CVF
  Conference on Computer Vision and Pattern Recognition (CVPR)}, 2021, pp.
  1164--1174.

\bibitem{undeepvo2018li}
R.~Li \emph{et~al.}, ``{UnDeepVO: Monocular Visual Odometry through
  Unsupervised Deep Learning},'' in \emph{International Conference on Robotics
  and Automation (ICRA)}.\hskip 1em plus 0.5em minus 0.4em\relax IEEE, 2018,
  pp. 7286--7291.

\bibitem{deep2020jau}
Y.-Y. Jau \emph{et~al.}, ``{Deep Keypoint-Based Camera Pose Estimation with
  Geometric Constraints},'' in \emph{IEEE/RSJ International Conference on
  Intelligent Robots and Systems (IROS)}, 2020, pp. 4950--4957.

\bibitem{better2020zhao}
W.~Zhao \emph{et~al.}, ``{Towards Better Generalization: Joint Depth-Pose
  Learning without PoseNet},'' in \emph{Proceedings of the IEEE/CVF Conference
  on Computer Vision and Pattern Recognition (CVPR)}, 2020, pp. 9151--9161.

\bibitem{unsuperpoint2019christiansen}
P.~H. Christiansen \emph{et~al.}, ``{UnsuperPoint: End-to-end Unsupervised
  Interest Point Detector and Descriptor},'' \emph{CoRR}, 2019.

\bibitem{selfsupervised2021tang}
J.~Tang \emph{et~al.}, ``{Self-Supervised 3D Keypoint Learning for Ego-Motion
  Estimation},'' in \emph{Proceedings of the Conference on Robot Learning
  (CoRL)}.\hskip 1em plus 0.5em minus 0.4em\relax PMLR, 2021, pp. 2085--2103.

\bibitem{unsupervised2021bian}
J.-W. Bian \emph{et~al.}, ``{Unsupervised Scale-Consistent Depth Learning from
  Video},'' \emph{International Journal of Computer Vision}, vol. 129, no.~9,
  pp. 2548--2564, 2021.

\bibitem{liu2023towards}
J.~Liu \emph{et~al.}, ``{Towards Better Data Exploitation in Self-Supervised
  Monocular Depth Estimation},'' \emph{IEEE Robotics and Automation Letters},
  vol.~9, no.~1, pp. 763--770, 2023.

\bibitem{planedepth2023wang}
R.~Wang \emph{et~al.}, ``{PlaneDepth: Self-Supervised Depth Estimation via
  Orthogonal Planes},'' in \emph{Proceedings of the IEEE/CVF Conference on
  Computer Vision and Pattern Recognition (CVPR)}, 2023, pp. 21\,425--21\,434.

\bibitem{he2016deep}
K.~He \emph{et~al.}, ``{Deep Residual Learning for Image Recognition},'' in
  \emph{Proceedings of the IEEE/CVF Conference on Computer Vision and Pattern
  Recognition (CVPR)}, 2016, pp. 770--778.

\bibitem{horn1981determining}
B.~K. Horn and B.~G. Schunck, ``{Determining Optical Flow},'' \emph{Artificial
  Intelligence}, vol.~17, no. 1-3, pp. 185--203, 1981.

\bibitem{deep2017zhu}
X.~Zhu \emph{et~al.}, ``{Deep Feature Flow for Video Recognition},'' in
  \emph{Proceedings of the IEEE/CVF Conference on Computer Vision and Pattern
  Recognition (CVPR)}, 2017, pp. 2349--2358.

\bibitem{gradient2010chapelle}
O.~Chapelle and M.~Wu, ``{Gradient Descent Optimization of Smoothed Information
  Retrieval Metrics},'' \emph{Information Retrieval}, vol.~13, no.~3, pp.
  216--235, 2010.

\bibitem{geiger2013vision}
A.~Geiger \emph{et~al.}, ``{Vision Meets Robotics: the KITTI Dataset},''
  \emph{The International Journal of Robotics Research}, vol.~32, no.~11, pp.
  1231--1237, 2013.

\bibitem{sun2023sc}
L.~Sun \emph{et~al.}, ``{SC-DepthV3: Robust Self-Supervised Monocular Depth
  Estimation for Dynamic Scenes},'' \emph{IEEE Transactions on Pattern Analysis
  and Machine Intelligence}, 2023.

\bibitem{hrdepth2021lyu}
X.~Lyu \emph{et~al.}, ``{HR-Depth: High Resolution Self-Supervised Monocular
  Depth Estimation},'' in \emph{Proceedings of the AAAI Conference on
  Artificial Intelligence (AAAI)}, vol.~35, no.~3, 2021, pp. 2294--2301.

\bibitem{selfsupervised2021zhou}
H.~Zhou \emph{et~al.}, ``{Self-Supervised Monocular Depth Estimation with
  Internal Feature Fusion},'' in \emph{Proceedings of the British Machine
  Vision Conference (BMVC)}, 2021.

\bibitem{monovit2022zhao}
C.~Zhao \emph{et~al.}, ``{MonoViT: Self-Supervised Monocular Depth Estimation
  with a Vision Transformer},'' in \emph{International Conference on 3D Vision
  (3DV)}, 2022, pp. 668--678.

\bibitem{swindepth2023shim}
D.~Shim and H.~J. Kim, ``{SwinDepth: Unsupervised Depth Estimation Using
  Monocular Sequences via Swin Transformer and Densely Cascaded Network},'' in
  \emph{International Conference on Robotics and Automation (ICRA)}.\hskip 1em
  plus 0.5em minus 0.4em\relax IEEE, 2023, pp. 4983--4990.

\bibitem{litemono2023zhang}
N.~Zhang \emph{et~al.}, ``{Lite-Mono: a Lightweight CNN and Transformer
  Architecture for Self-Supervised Monocular Depth Estimation},'' in
  \emph{Proceedings of the IEEE/CVF Conference on Computer Vision and Pattern
  Recognition (CVPR)}, 2023, pp. 18\,537--18\,546.

\bibitem{dynamodepth2023sun}
Y.~Sun and B.~Hariharan, ``{Dynamo-Depth: Fixing Unsupervised Depth Estimation
  for Dynamical Scenes},'' in \emph{Advances in Neural Information Processing
  Systems (NeurIPS)}, vol.~36, 2023.

\bibitem{disentangling2022feng}
Z.~Feng \emph{et~al.}, ``{Disentangling Object Motion and Occlusion for
  Unsupervised Multi-Frame Monocular Depth},'' in \emph{Proceedings of the
  European Conference on Computer Vision (ECCV)}.\hskip 1em plus 0.5em minus
  0.4em\relax Springer, 2022, pp. 228--244.

\bibitem{uhrig2017sparsity}
J.~Uhrig \emph{et~al.}, ``{Sparsity Invariant CNNs},'' in \emph{International
  Conference on 3D Vision (3DV)}.\hskip 1em plus 0.5em minus 0.4em\relax IEEE,
  2017, pp. 11--20.

\bibitem{yu2018deep}
F.~Yu \emph{et~al.}, ``{Deep Layer Aggregation},'' in \emph{Proceedings of the
  IEEE/CVF Conference on Computer Vision and Pattern Recognition (CVPR)}, 2018,
  pp. 2403--2412.

\bibitem{are2012geiger}
A.~Geiger \emph{et~al.}, ``{Are We Ready for Autonomous Driving? the KITTI
  Vision Benchmark Suite},'' in \emph{Proceedings of the IEEE/CVF Conference on
  Computer Vision and Pattern Recognition (CVPR)}, 2012, pp. 3354--3361.

\bibitem{make3d2009saxena}
A.~Saxena \emph{et~al.}, ``{Make3D: Learning 3D Scene Structure from a Single
  Still Image},'' \emph{IEEE Transactions on Pattern Analysis and Machine
  Intelligence}, vol.~31, no.~5, pp. 824--840, 2009.

\bibitem{selfsupervised2019chen}
Y.~Chen \emph{et~al.}, ``{Self-Supervised Learning with Geometric Constraints
  in Monocular Video: Connecting Flow, Depth, and Camera},'' in
  \emph{Proceedings of the IEEE/CVF International Conference on Computer Vision
  (ICCV)}, 2019, pp. 7063--7072.

\bibitem{shen2019beyond}
T.~Shen \emph{et~al.}, ``{Beyond Photometric Loss for Self-Supervised
  Ego-Motion Estimation},'' in \emph{International Conference on Robotics and
  Automation (ICRA)}.\hskip 1em plus 0.5em minus 0.4em\relax IEEE, 2019, pp.
  6359--6365.

\bibitem{sturm2012benchmark}
J.~Sturm \emph{et~al.}, ``{A Benchmark for the Evaluation of RGB-D SLAM
  Systems},'' in \emph{IEEE/RSJ International Conference on Intelligent Robots
  and Systems (IROS)}.\hskip 1em plus 0.5em minus 0.4em\relax IEEE, 2012, pp.
  573--580.

\bibitem{kingma2014adam}
D.~P. Kingma and J.~Ba, ``{Adam: a Method for Stochastic Optimization},'' in
  \emph{International Conference on Machine Learning (ICML)}, 2015.

\bibitem{yan2021channel}
J.~Yan \emph{et~al.}, ``{Channel-Wise Attention-Based Network for
  Self-Supervised Monocular Depth Estimation},'' in \emph{International
  Conference on 3D Vision (3DV)}.\hskip 1em plus 0.5em minus 0.4em\relax IEEE,
  2021, pp. 464--473.

\end{thebibliography}

\end{document}